\documentclass[sigconf, nonacm]{acmart}
\AtBeginDocument{}
\newcommand\projectname{RadarTwin}
\usepackage{pifont}
\definecolor{okgreen}{RGB}{0,140,55}
\definecolor{nored}{RGB}{200,30,30}
\definecolor{amberp}{RGB}{220,140,0}
\newcommand{\cmark}{\textcolor{okgreen}{\ding{51}}}      
\newcommand{\xmark}{\textcolor{nored}{\ding{55}}}        
\newcommand{\pmark}{\textcolor{amberp}{\textbf{\textasciitilde}}} 
\begin{document}

\title{RadarTwin: Scene-Specific mmWave Radar Simulation and Learning for Mobile Indoor Perception}

\author{Emily Bejerano}
\affiliation{%
  \institution{Columbia University}
  \city{New York}
  \state{NY}
  \country{USA}}
\email{eg3205@columbia.edu}

\author{Federico Tondolo}
\affiliation{%
  \institution{Columbia University}
  \city{New York}
  \state{NY}
  \country{USA}}
\email{ft2505@columbia.edu}

\author{Devang Gupta}
\affiliation{%
  \institution{Columbia University}
  \city{New York}
  \state{NY}
  \country{USA}}
\email{dg3529@columbia.edu}

\author{Aaron Mano Cherian}
\affiliation{%
  \institution{Columbia University}
  \city{New York}
  \state{NY}
  \country{USA}}
\email{amc2535@columbia.edu}

\author{Taeyoo Kim}
\affiliation{%
  \institution{Columbia University}
  \city{New York}
  \state{NY}
  \country{USA}}
\email{tk3151@columbia.edu}

\author{Ayaan Qayyum}
\affiliation{%
  \institution{Columbia University}
  \city{New York}
  \state{NY}
  \country{USA}}
\email{aaq2109@columbia.edu}

\author{Xiaofan Yu}
\affiliation{%
  \institution{University of California, Merced}
  \city{Merced}
  \state{CA}
  \country{USA}}
\email{xiaofanyu@ucmerced.edu}

\author{Xiaofan Jiang}
\affiliation{%
  \institution{Columbia University}
  \city{New York}
  \state{NY}
  \country{USA}}
\email{jiang@ee.columbia.edu}

\renewcommand{\shortauthors}{Bejerano et al.}

\begin{abstract}
Millimeter-wave (mmWave) radar perception is limited by data scarcity: models trained on existing radar datasets fail to generalize to new objects, environments, and sensing trajectories. We present RadarTwin, a framework for generating deployment-specific radar training data before real data collection. Given a 3D reconstruction of a target space (phone LiDAR, robot-mounted sensing, or RGB-to-3D), RadarTwin uses a vision-language model to infer radar-relevant surface materials and a physics-based ray tracer to synthesize raw frequency-modulated continuous-wave (FMCW) radar measurements with multi-bounce propagation. To study what transfers from simulation to reality, we collect a paired real-simulated dataset spanning household objects, material classes, distances, rotations, translations, and mobile sensing trajectories. We show that simulated and real radar share the same object-discriminative shape and material features, and that modeling the environment's multipath is essential to matching real measurements. A representation trained on simulation alone recognizes real objects at $2.5\times$ chance with no real radar labels, and a few labeled examples raise this to 95.3\% on a 12-way recognition task. RadarTwin enables training radar perception for a new space before any real radar data is collected there.
\end{abstract}

\maketitle
\raggedbottom

\begin{figure}[t!]
    \centering
    \includegraphics[width=0.9\columnwidth]{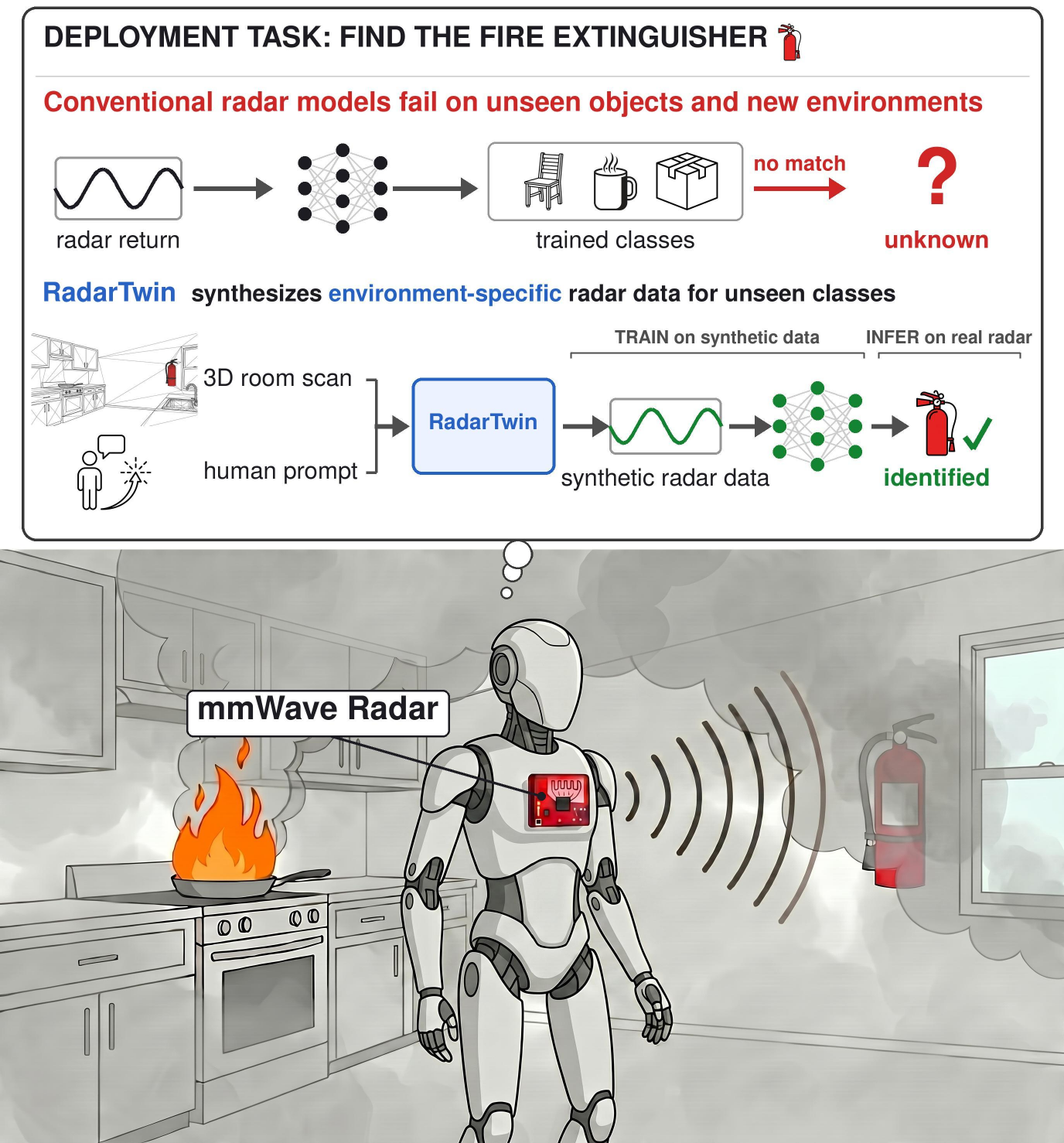}
    \vspace{-0.8em}
    \caption{\textbf{Motivating example.}
    A mobile robot may need to detect a safety-critical target, such as a fire extinguisher, in a smoke-filled environment. Real radar datasets provide limited coverage of objects, scenes, viewpoints, radar hardware, and chirp configurations. \projectname{} bridges this gap by generating deployment-specific radar training data for unseen objects and sensing configurations.}
    \vspace{-1.0em}
    \Description{A robot searches for a fire extinguisher in a smoky room. The lower panel contrasts existing radar datasets with deployment needs, including novel objects, new environments, diverse robot viewpoints, different radar hardware, and different chirp configurations.}
    \label{fig:motivation}
\end{figure}

\section{Introduction}

Millimeter-wave (mmWave) radar can sense indoor environments in conditions where optical sensors degrade, including smoke, dust, fog, and darkness. It also captures spatial and material structure without recording visually identifiable imagery, making it attractive for safety-critical, robotic, and privacy-sensitive deployments~\cite{lu2020millimap, lam2025minav, he2023fusang}. Yet deployment remains limited by training data. Indoor radar datasets are far smaller than visual datasets, and radar signatures are highly sensitive to object geometry, material composition, sensing trajectory, and room-specific multipath. As a result, a model trained in one environment may not generalize to another, and collecting labeled radar data for every new space, object set, and trajectory is often impractical.

Simulation offers a way to generate deployment-specific radar data before real data collection, but only if the simulator captures the factors that dominate indoor radar returns. Geometry alone is insufficient. Two objects or surfaces with similar shape can scatter very differently if one is metal and the other is wood, plastic, ceramic, or glass. Indoor radar measurements are also shaped not only by the target object, but by reflections from walls, floors, furniture, and fixtures. Existing radar simulators and cross-modal generators make important progress, but many require manually authored CAD scenes and material labels, synthesize scenes without correspondence to the deployment space, assume simplified or uniform materials, or focus on human bodies rather than object-centric indoor perception~\cite{chen2023rfgenesis, ahuja2021vid2doppler, ling2024uranus}. These limitations make it difficult to generate training data that reflects a specific indoor deployment.

The timing is favorable for a different approach. Robotics, embodied AI, augmented reality, and 3D vision have already produced a large ecosystem of indoor scans, object meshes, scene assets, and reconstruction tools. Commodity devices can now capture usable 3D geometry, and many robotic platforms already maintain geometric maps for navigation and planning. These assets are usually built for visual simulation or robot control rather than radar, but they provide exactly the spatial scaffold that radar simulation needs. If these existing 3D resources can be augmented with electromagnetic material properties, radar training data can be generated with little manual scene authoring and without routine site-specific radar data collection as simulator input.

We present \projectname{}, a framework for generating physics-based radar training data from a reconstructed indoor scene. Given a 3D reconstruction of a target space, obtained for example from phone LiDAR, robot-mounted RGB-D sensing, existing robotics assets, or RGB-to-3D reconstruction, \projectname{} segments the scene and uses a vision-language model (VLM) to infer radar-relevant surface materials. These material labels are mapped to electromagnetic properties and attached to the scene geometry. A ray tracer then synthesizes raw FMCW radar measurements with multi-bounce propagation under specified object placements and sensor motion. This produces radar data that is tied to the geometry and material composition of the intended deployment space, while requiring minimal manual recording or annotation effort from the user.

This paper studies what such simulation can and cannot transfer to real radar deployments. We collect a paired real-simulated dataset spanning household objects, material classes, distances, rotations, translations, and mobile sensing trajectories. Using this dataset, we disentangle the sim-to-real gap across shape, material, size, and environment factors. The results show that \projectname{} preserves distance-invariant shape signatures and material-class structure, while fine size differences are limited by radar resolution and environment-specific multipath is the primary source of the remaining sim-to-real gap. Building on these findings, we learn a radar representation aligned with physical text descriptions of object geometry and material, and evaluate simulation-trained recognition on real radar measurements across supervision levels. Because simulated and real radar map onto the same object-discriminative features in this representation, a model trained on simulation alone recognizes real objects with no real radar labels, and a few labeled examples lift accuracy further.

\noindent\textbf{This paper makes the following contributions:}

\begin{itemize}
\setlength{\topsep}{1pt}
\setlength{\partopsep}{0pt}
\setlength{\itemsep}{2pt}
\setlength{\parsep}{0pt}
\setlength{\parskip}{0pt}

\item An end-to-end framework that generates physically grounded mmWave
radar training data from commodity 3D scans by inferring per-surface
electromagnetic material properties with VLM reasoning and synthesizing
raw FMCW radar measurements through multi-bounce ray tracing.

\item A paired real-simulated evaluation dataset spanning household objects,
material classes, distances, rotations, translations, and mobile sensing
trajectories, so that what transfers from simulation to real radar can be
analyzed directly.

\item A disentangled characterization of the sim-to-real gap across shape,
material, size, and environment factors. We show that simulation preserves
distance-invariant shape signatures and material-class structure, while
fine size differences are limited by radar resolution and
environment-specific multipath is the primary source of the remaining
sim-to-real gap.

\item A domain-invariant radar representation aligned with physical
descriptions of object geometry and material, through which
simulation-trained features transfer to real radar measurements for
object and material recognition.

\item A scene-level evaluation on mobile indoor object recognition across
three supervision levels. The results show that material-aware scene
reconstruction and physics-based simulation provide useful training priors
for unseen objects and deployment environments.

\end{itemize}

\begin{table*}[t]
  \centering
  \caption{Comparison of RF/radar simulators and cross-modal data generators
  for learning-based perception. \cmark/\xmark\ denote a capability present or absent
  and \pmark\ denotes partial support. ``-'' marks an axis that does not apply because
  purely learned generators have no explicit scene, material, or propagation model.
  Multipath denotes multi-bounce environment modeling, with bounce count shown for ours.
  Real scene denotes whether the method models the actual deployment space rather than
  a synthetic or absent one. No RF data denotes whether the method requires no real radar
  measurements as input. \projectname{} (bold) is the only method that infers per-surface
  materials automatically from an available 3D scene representation while modeling
  environment multipath and requiring no RF data.}
  \label{tab:positioning}
  \footnotesize
  \setlength{\tabcolsep}{4.5pt}
  \renewcommand{\arraystretch}{1.18}
  \begin{tabular}{l l l l c c c l}
    \toprule
    Method & Mechanism & Scene input & Material model &
    Multipath & \shortstack{Real\\scene} & \shortstack{No RF\\data} & Target \\
    \midrule
    EM solvers~\cite{matlab_radar,ansys_hfss,remcom}        & full-wave / RT     & manual CAD            & manual per-surf.        & \cmark        & \cmark & \cmark & general \\
    Sionna~RT~\cite{hoydis2023sionna}                       & diff.\ ray trace   & imported mesh         & ITU per-surf.           & \cmark        & \cmark & \cmark & comms channel \\
    ViRa~\cite{schoffmann2021vira}                          & game engine        & engine mesh           & simplified              & \pmark        & \cmark & \cmark & robotics \\
    Shenron/C-Shenron~\cite{shenron2024,mishra2025cshenron} & ray tracing        & LiDAR\,+\,cam.        & reflectivity            & \cmark        & \cmark & \cmark & driving \\
    Vid2Doppler~\cite{ahuja2021vid2doppler}                 & video projection   & video                 & none                    & \xmark        & \xmark & \cmark & human \\
    RF-Genesis~\cite{chen2023rfgenesis}                     & RT\,+\,diffusion   & vision\,+\,diffusion  & uniform                 & \cmark        & \xmark & \cmark & human \\
    RF-Diffusion~\cite{chi2024rfdiffusion}                  & diffusion          & -                     & -                       & -             & \xmark & \xmark & signal gen. \\
    mmCLIP~\cite{cao2024mmclip}                             & mocap synthesis    & human mesh            & none                    & \xmark        & \xmark & \cmark & activity \\
    RFCanvas~\cite{chen2024rfcanvas}                        & RT\,+\,learned     & mono.\,+\,few-shot RF & learned                 & \cmark        & \cmark & \xmark & channel \\
    \midrule
    \textbf{\projectname{} (Ours)} & \textbf{ray tracing} & \textbf{3D scene repr.} & \textbf{VLM\,$\rightarrow$\,ITU} & \textbf{\cmark\,(4-bounce)} & \textbf{\cmark} & \textbf{\cmark} & \textbf{objects \& materials} \\
    \bottomrule
  \end{tabular}
\end{table*}\section{Background and Related Work}
\subsection{mmWave Radar Sensing Primer}

Frequency-modulated continuous-wave (FMCW) radar measures distance by transmitting a radio-frequency sweep, or chirp, whose frequency increases linearly over time. Reflections from objects return after a short delay. Mixing the delayed echo with the transmitted chirp produces a beat frequency proportional to round-trip travel time, and therefore to range. A Fast Fourier Transform (FFT) within each chirp converts the signal into a range profile.

A radar frame contains multiple chirps. Phase changes across chirps reveal Doppler velocity, i.e., whether a target is moving toward or away from the radar. A second FFT across chirps estimates this motion. Multiple-input multiple-output (MIMO) antennas estimate angle from phase differences across the antenna array. After range and Doppler processing, constant false-alarm-rate (CFAR) detection removes weak background responses, leaving a sparse point cloud of range, velocity, and angle.

\projectname{} synthesizes raw FMCW measurements and applies the same processing pipeline to simulated and recorded data. This lets us compare simulation and reality at the radar signal level, before a downstream learning model introduces task-specific transformations.

Radar returns depend on both geometry and material. A surface's relative permittivity $\epsilon_r$ and conductivity $\sigma$ determine how much energy it reflects: metals usually reflect strongly, while wood, plastic, ceramic, and glass return material-dependent fractions. ITU-R P.2040-4~\cite{itu2040} tabulates these properties for common building materials up to 100~GHz.

Geometry alone is therefore insufficient for indoor radar simulation. A metal door and a wooden door with the same shape can produce different returns, and indoor rooms add multipath from walls, floors, furniture, and fixtures. A simulator that ignores per-surface materials, or assigns one material to the whole room, loses both object-level reflectivity differences and room-specific multipath structure~\cite{lu2020millimap, he2023fusang, dodds2025nlos}.

\subsection{Radar Simulation and Cross-Modal Generation}

Radar simulation has been studied from several directions. Commercial EM tools,
including MATLAB Radar Toolbox~\cite{matlab_radar}, Ansys HFSS
SBR+~\cite{ansys_hfss}, and Remcom~\cite{remcom}, can model RF propagation with
high fidelity, but they typically require expert-built CAD scenes and manually
assigned material properties. Other simulators are designed for specific domains.
For example, Shenron~\cite{shenron2024} and C-Shenron~\cite{mishra2025cshenron}
use LiDAR and camera data to support material-aware simulation for driving
scenarios, while Sionna~RT~\cite{hoydis2023sionna} provides differentiable ray
tracing for wireless communication channels, and
RFCanvas~\cite{chen2024rfcanvas} fits per-surface channel properties from a
few real RF measurements, which presupposes RF hardware already deployed at
the target site.

A separate line of work generates radar or RF measurements from other sensing
modalities. Vid2Doppler~\cite{ahuja2021vid2doppler} maps video to Doppler
signatures, RF-Genesis~\cite{chen2023rfgenesis} combines ray tracing with
diffusion-generated scenes and human meshes, RF-Diffusion~\cite{chi2024rfdiffusion}
generates time-frequency radar representations, and Uranus~\cite{ling2024uranus}
targets gesture sensing. These methods show that cross-modal generation can
produce useful RF-like signals, but they are usually focused on humans or
gestures and do not model the material composition and multipath structure of a
specific indoor deployment.

\projectname{} targets this setting. Given an available 3D scene representation of the deployment
space, it reconstructs the room, infers per-surface electromagnetic material
properties from visual context, and simulates raw FMCW radar measurements with
multi-bounce propagation. Our implementation uses Mitsuba~3~\cite{jakob2022mitsuba3}
and adapts the RF-Genesis ray-tracing pipeline~\cite{chen2023rfgenesis} for
object-centric and room-scale indoor radar simulation. Table~\ref{tab:positioning}
compares \projectname{} with prior simulators and generators. As the table makes explicit, no prior system occupies our setting:
\projectname{} is the only approach that starts from the actual indoor
scene, assigns radar-relevant materials without manual labeling, models
environment multipath, and requires no real RF measurements to generate
training data.

\subsection{Indoor mmWave Perception and VLMs for Materials}
mmWave radar is used for indoor mapping through smoke~\cite{lu2020millimap},
SLAM~\cite{lu2024radarize}, navigation~\cite{lam2025minav}, and object
recognition~\cite{he2023fusang, dodds2025nlos}. Public indoor radar datasets remain scarce. Fusang~\cite{he2023fusang}
and the Indoor FireRescue dataset~\cite{duan2025ifr} provide indoor
recordings, but neither pairs them with matched scene-specific
simulations, which our evaluation dataset provides. The closest prior work to our
contrastive component is mmCLIP~\cite{cao2024mmclip}, which pretrains on
synthetic mmWave (synthesized from human motion-capture sequences) aligned
to LLM-generated text and transfers zero-shot to real data for activity
recognition. We share that synthetic-pretrain, text-alignment, sim-to-real
paradigm, but differ fundamentally. Our synthetic data is produced by
physics-based ray tracing of a material-labeled, VLM-reconstructed
deployment scene, rather than motion-capture synthesis with no material or
environment model, and we target indoor object and material recognition
rather than human activity (Section~\ref{sec:contrastive}). A recurring challenge
across indoor radar perception is the need for environment-specific data. For materials, texture-based vision is
insufficient. An industrial door is metal due to code, not appearance. VLMs
combine visual recognition with world knowledge.
InternVL2.5~\cite{chen2024internvl} performs well on physical-property
reasoning~\cite{chow2025physbench}. We use a VLM to classify each surface into
materials drawn from MINC~\cite{bell2015minc}, each mapped to ITU-R~P.2040
electromagnetic properties~\cite{itu2040}.

\begin{figure*}[t]
    \centering
    \includegraphics[width=0.92\textwidth]{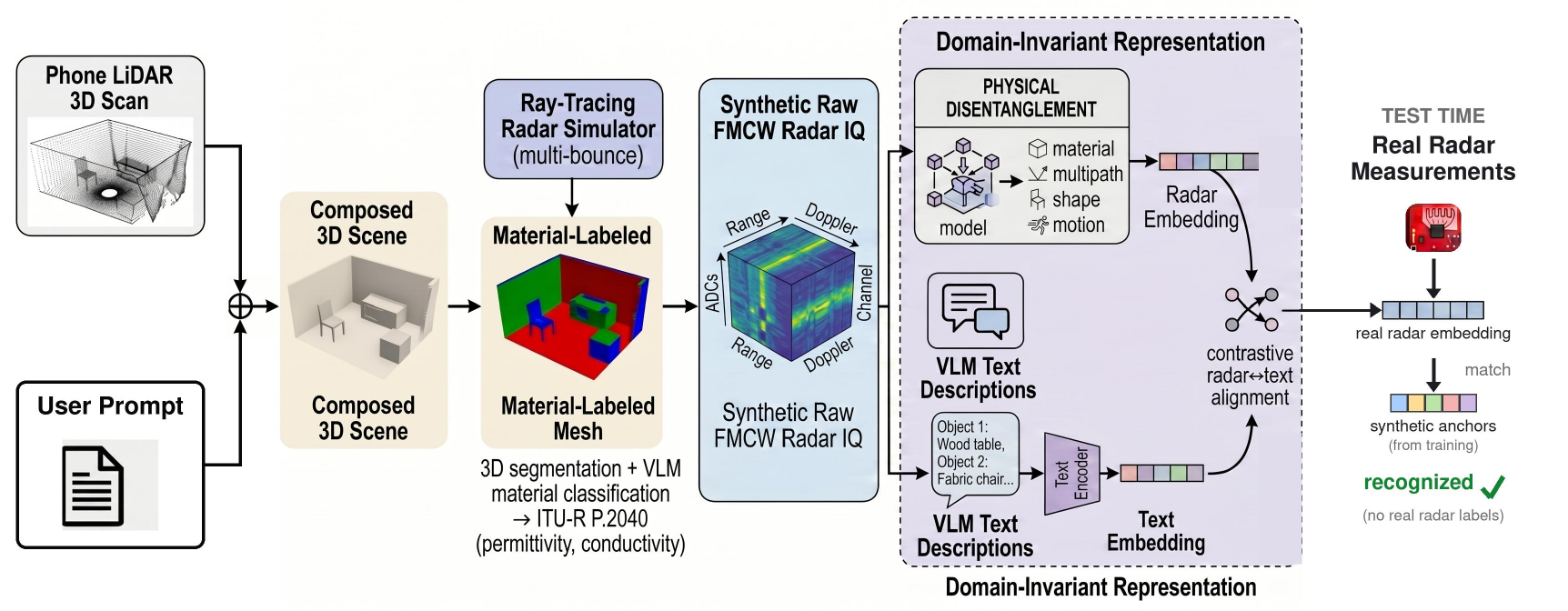}
    \vspace{-0.8em}
    \caption{\textbf{\projectname{} system overview.}
A 3D reconstruction and user-specified object placement are composed into a material-labeled scene. A VLM assigns per-surface electromagnetic materials, a ray tracer simulates multi-bounce FMCW returns, and a contrastive representation aligns simulated radar with physical text descriptions for transfer to real measurements.}
    \vspace{-1.0em}
    \Description{Overview of the \projectname{} pipeline. Inputs include a phone LiDAR scan, 3D object models, and a user prompt. The system composes a 3D scene, assigns material labels using segmentation and VLM material reasoning, simulates synthetic raw FMCW radar data with a ray-tracing radar simulator, and learns a domain-invariant representation aligned with VLM text descriptions for transfer to real radar measurements.}
    \label{fig:pipeline}
\end{figure*}
\section{System Design}
\label{sec:frame}

\projectname{} transforms an available 3D scene representation into synthetic radar through three stages:
(1) material-aware scene reconstruction, (2) physics-based ray-tracing
simulation, and (3) FMCW signal processing. Target objects and the sensing
trajectory are placed in the reconstructed scene either from explicit
coordinates or from a natural-language prompt that a local LLM parses into
a scene layout. The evaluation in this paper uses measured placements so
that simulations match the real recordings. The stages are decoupled, so a
reconstructed scene can be reused across different radar configurations
(Fig.~\ref{fig:pipeline}).

\subsection{Material-Aware Scene Reconstruction}

The reconstruction stage turns a commodity 3D reconstruction into a material-labeled mesh suitable for electromagnetic ray tracing. The input can come from several sources, including a phone LiDAR scan, robot-mounted RGB-D sensing, an existing 3D asset, or an RGB-only reconstruction pipeline. In our implementation, the preferred input is synchronized depth$+$RGB from a LiDAR-equipped phone; for devices without LiDAR, we support an RGB-only fallback using a monocular geometry model (MoGe~\cite{wang2025moge}). In all cases, the output is a triangulated mesh of the deployment space aligned with RGB views for material identification.

\begin{figure*}[t]
  \centering
  \includegraphics[width=0.78\textwidth]{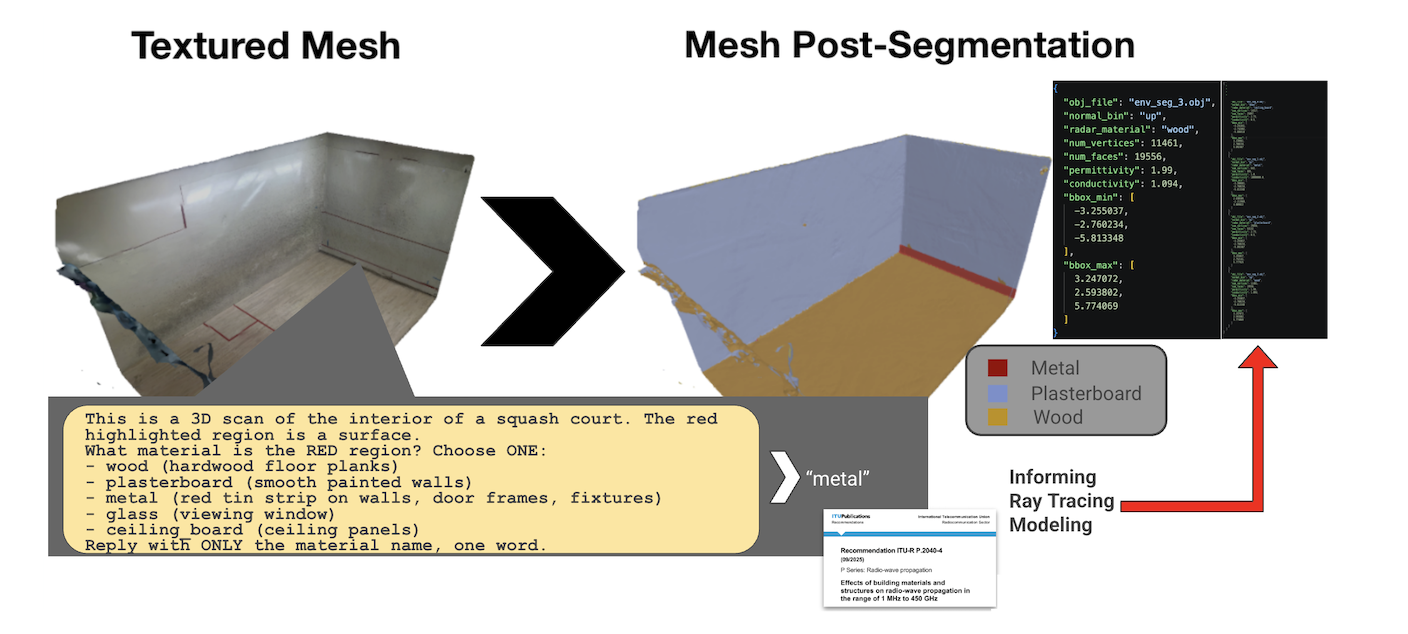}
  \caption{\textbf{Material-aware scene reconstruction} (squash court). The scanned
mesh (left) is segmented and each segment is classified by the VLM
(right). Inset: an example query, identifying a tin strip as metal. The
resulting manifest carries ITU-R P.2040 properties into the ray
tracer.}
  \label{fig:material_seg}
\end{figure*}

The key challenge is that radar reflectivity depends on electromagnetic material properties, not geometry alone. A painted metal cabinet and a painted wooden cabinet may look similar but produce different radar signatures. RadarTwin addresses this in two steps. First, a 3D segmentation process partitions the scene into coherent surfaces, such as walls, floors, furniture facets, and object parts. Second, a vision-language model (InternVL2.5-8B) inspects each segment in its visual context and assigns a material label from the 23~MINC categories~\cite{bell2015minc} (Fig.~\ref{fig:material_seg}). The VLM uses both appearance and world knowledge, for example inferring that a fire door is likely sheet metal despite its paint. This matters because purely appearance-based classifiers can conflate visually similar materials that scatter radar energy very differently~\cite{chow2025physbench}.

Each of the MINC labels is then mapped to corresponding electromagnetic properties through the
ITU-R~P.2040 recommendation~\cite{itu2040}, which tabulates frequency-dependent
relative permittivity $\epsilon_r$ and conductivity $\sigma$ for common
materials. These values set the Fresnel reflection coefficients the ray tracer
requires. If the VLM identifies a material not covered by the ITU, the mask in question is mapped to
its electromagnetically closest tabulated entry rather than an unclassified label. For instance, ``porcelain'' would inherit glass-like permittivity
($\epsilon_r\!\approx\!6$) rather than defaulting to an unrelated baseline value.
The output is a 3D environment in which every polygon face carries a material label. Because the scene is self-contained, it can be re-rendered across a range of radar configurations.

Two components are deliberately modular: the specific pipeline used for
3D segmentation and the exact VLM input format (cropped RGB frames vs.\
rendered mesh views). Our validation establishes whether the simulator, given
a material-labeled scene, produces realistic radar signatures, and the
downstream cost of imperfect material assignment is bounded directly by the
reconstruction ablation (Section~\ref{sec:rover-recog}): replacing every VLM
label with a single uniform material costs $0.035$ of label-free
recognition accuracy, while removing the environment costs $0.089$. The
upstream perception modules can therefore be treated as interchangeable.

\subsection{Physics-Based Radar Simulation}

The simulation stage synthesizes the raw FMCW signal that the radar would measure in the reconstructed scene. We build on a Mitsuba~3~\cite{jakob2022mitsuba3} ray-tracing pipeline~\cite{chen2023rfgenesis}, treating the transmit antenna as a source of rays that propagate through the scene, reflect from surfaces, and return to the receive array. At each intersection, the surface's ITU material properties determine its reflection strength. Conductive surfaces reflect strongly, while dielectric materials return a material-dependent fraction of the incident energy. The ray tracer also accounts for object geometry, viewing aspect, and surface roughness, which broadens otherwise specular reflections.

We trace up to four bounces, so each simulated frame contains both direct object reflections and room-specific multipath from walls, floors, furniture, and interactions between objects and the environment. This is essential indoors, where indirect paths can dominate the measured radar response. Material-brightness comparisons in our evaluation are therefore made only at matched geometry and aspect, isolating the effect of material from pose-dependent scattering.

The simulated scene is assembled to match the real measurement geometry. Objects are placed at their measured stand-offs, and antenna phase centers are positioned at the corresponding sensor locations so that simulated and real ranges align. Motion is applied per frame for turntable rotations, linear-rail translations, and recorded rover trajectories. For every pose, the returned paths are converted into an FMCW intermediate-frequency signal. Each path contributes a complex sinusoid whose beat frequency encodes range, while its phase across the MIMO transmit-receive pairs preserves angle.

We output simulated measurements in the raw ADC format of the TI IWR1843BOOST, using the same carrier frequency, bandwidth, chirp slope, samples per chirp, and chirps per frame as the real radar. The same downstream processing pipeline, including range and Doppler FFTs followed by CFAR detection, is then applied to both simulated and recorded data, producing comparable radar point clouds. Because scene reconstruction, propagation, and signal synthesis are decoupled, the same material-labeled room can be re-rendered for different radar configurations, stand-offs, and motion patterns without re-scanning the space.

\section{Implementation}
\label{sec:impl}

We implement \projectname{} using a TI IWR1843BOOST 77\,GHz FMCW radar and a physics-based simulator built on Mitsuba~3 and RF-Genesis. For deployment-realistic evaluation, the radar is mounted on a ROS-controlled mecanum-wheel robot capable of omnidirectional motion, which supports the forward, lateral, diagonal, and rotational trajectories common in indoor navigation. The same radar configuration is used across all experiments (turntable, linear rail, mobile robot).

\subsection{Radar Hardware}
Raw IQ data are captured through a DCA1000EVM and processed by a common pipeline (range FFT, Doppler FFT, CFAR detection, feature extraction) applied identically to real and simulated recordings (Table~\ref{tab:radar_config}). Holding radar parameters, antenna geometry, chirp configuration, and processing identical across domains isolates scene-reconstruction and propagation effects from hardware-induced differences.
\begin{table}[t]
\centering
\caption{Radar configuration used throughout all experiments.}
\label{tab:radar_config}
\begin{tabular}{lr}
\toprule
Parameter & Value \\
\midrule
Carrier frequency & 77 GHz \\
Chirp slope & 70 MHz/$\mu$s \\
ADC sample rate & 5.21 Msps \\
Sampled bandwidth & 3.44 GHz \\
Range resolution & 0.044 m \\
ADC samples/chirp & 256 \\
Chirps/frame & 16 \\
TX antennas & 2 \\
RX antennas & 4 \\
Virtual array size & 8 \\
Frame rate & 10 Hz \\
\bottomrule
\end{tabular}
\end{table}
\subsection{Simulator}
The simulator builds on Mitsuba~3~\cite{jakob2022mitsuba3}
(\texttt{cuda\_ad\_rgb} variant, CUDA backend) and the RF-Genesis ray-tracing
pipeline~\cite{chen2023rfgenesis}, with material physics replaced to follow
ITU-R~P.2040-4~\cite{itu2040} on a per-surface basis. Each segmented surface
in the reconstructed mesh is assigned a complex permittivity $\epsilon_r$ and
conductivity $\sigma$ from the ITU material table at 77~GHz, plus a
Rayleigh-roughness parameter calibrated from the published microstructure of
each MINC class (e.g., metal $\sigma_h = 0.1$~mm, rougher dielectrics
0.3--3~mm). When the VLM names a material the ITU table does not tabulate
explicitly, we map it to the EM-closest tabulated material (e.g., porcelain
$\rightarrow$ glass, $\epsilon_r\!\approx\!6$) rather than a generic default,
which prevents silent material misassignment of high-permittivity surfaces.
Rays are cast once from the sensor position and Fresnel reflection is
evaluated at every ray-surface intersection, recursing up to four bounces.
The four-bounce limit is validated empirically rather than by a fixed
per-bounce budget. It suffices to reproduce the off-target multipath
fraction observed in real recordings (Section~\ref{sec:env-ablation}), though
metal-rich rooms could in principle sustain energy beyond four bounces. The returned set of paths is converted to FMCW IQ
matched to the IWR1843BOOST chirp schedule by accumulating per-path complex
exponentials whose beat frequency encodes round-trip delay and whose
chirp-to-chirp phase progression encodes radial velocity. From the single traced path set, each TX-RX pair's IQ is synthesized by
applying the per-element phase of the virtual-array manifold under a
far-field approximation. Each scatterer contributes its array-geometry
phase offset per virtual channel, so the angular response is preserved
analytically rather than by multistatic tracing. Antenna phase centers
follow the sensor position per frame so that range is reported in absolute
coordinates rather
than relative to a fixed origin (Section~\ref{sec:rail}). The output is therefore
bit-compatible with the DCA1000EVM dump format, and the same offline
processing pipeline applies to both domains. Simulation and representation
training run offline on a single NVIDIA RTX~5090 GPU. Onboard capture runs
on the platform's NVIDIA Jetson Orin Nano (Section~\ref{sec:rover-platform}).

\subsection{Mobile Robot Platform}
\label{sec:rover-platform}
\begin{figure}[t]
\centering
\includegraphics[width=0.38\linewidth]{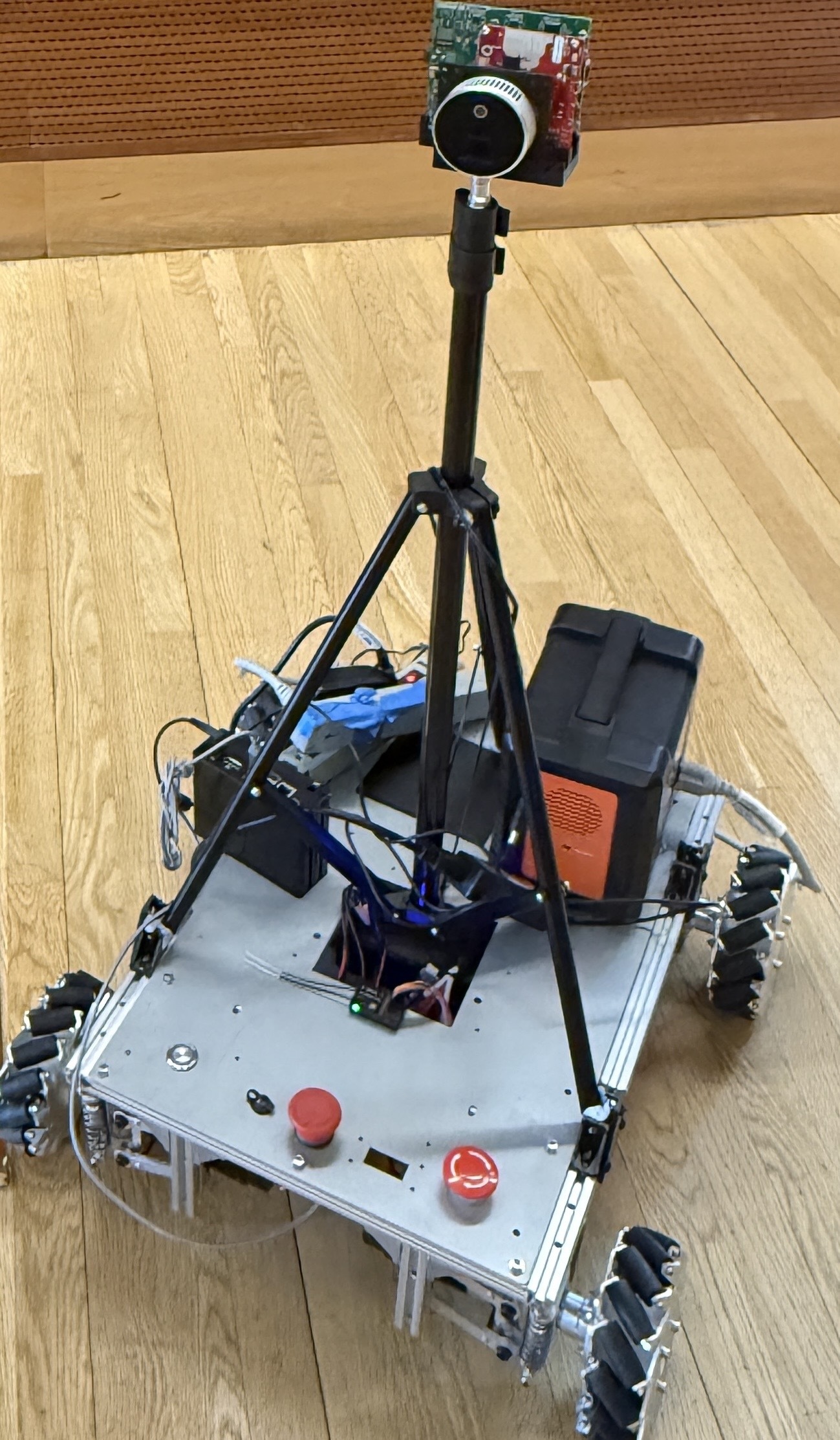}
\caption{The mobile platform: IWR1843BOOST mmWave radar and
time-synchronized RGB-D camera mast-mounted on a mecanum-wheel base.}
\label{fig:rover_platform}
\end{figure}

The scene-level evaluation (Section~\ref{sec:rover}) is conducted on an NVIDIA Jetson Orin
Nano-powered ROS-based mobile robot platform (Fig.~\ref{fig:rover_platform}) with an omnidirectional
(omni-wheeled) base and integrated multimodal USB sensors for real-time
perception and sensor fusion. The radar is rigidly mounted at $\sim$0.4~m
height, a stand-off chosen to clear typical floor clutter while keeping
boresight intersecting common indoor-object centers (cans, mugs, bottles)
across the $0.5$--$2.0$~m stand-off range we evaluate. The omnidirectional
drive matters. A differential-drive base would couple any lateral motion to a
yaw rotation about its center, which would change aspect angle and complicate
sim-to-real alignment. Omnidirectional drive decouples translation from
rotation, so a commanded pure-lateral sweep produces a pure-lateral
trajectory and a commanded pure-depth approach produces a pure-radial
trajectory. This decoupling lets us simulate each motion regime cleanly using
the trajectory descriptor the radar geometry exposes (translation along a
single axis), and lets us tag each recording with a single ground-truth motion
type that maps one-to-one onto a simulated trajectory. The platform also
carries an RGB-D camera time-synchronized to the radar capture, used here
only for trajectory matching in the qualitative comparisons. Perception relies on
the radar alone.

We validate what the simulator captures through controlled microbenchmarks, a
turntable (rotation) and a precision linear rail (translation), isolating
specific physical capabilities one at a time. The deployment-realistic
downstream evaluation is then performed with a mobile (rover) platform in
Section~\ref{sec:rover}.

\noindent\textbf{Fidelity at the feature level.}
A simulator built to generate training data need not reproduce the real
signal sample-for-sample. At 77~GHz, with hardware noise and unmodeled
micro-structure, it cannot. Raw range-Doppler returns differ substantially between
simulation and reality, so a pixel-level signal comparison understates a
simulator that is in fact useful. The question that matters is not whether the
signal matches point-for-point, but whether the features that
distinguish objects (shape signature, material class, motion) stand in the
same relationship to one another in simulation as in reality, because those
features, not the raw waveform, are what a perception model consumes. We
therefore measure fidelity at the feature level. We show that the inter-object
structure these features induce is preserved from simulation to real
(high rank and structure-preservation correlation) even where absolute signals
are not. We establish the relationship structurally here, and confirm it
transfers through recognition (Section~\ref{sec:recog}) and at scene scale
(Section~\ref{sec:rover}).

\begin{figure*}[t]
  \centering
  \includegraphics[width=0.56\textwidth]{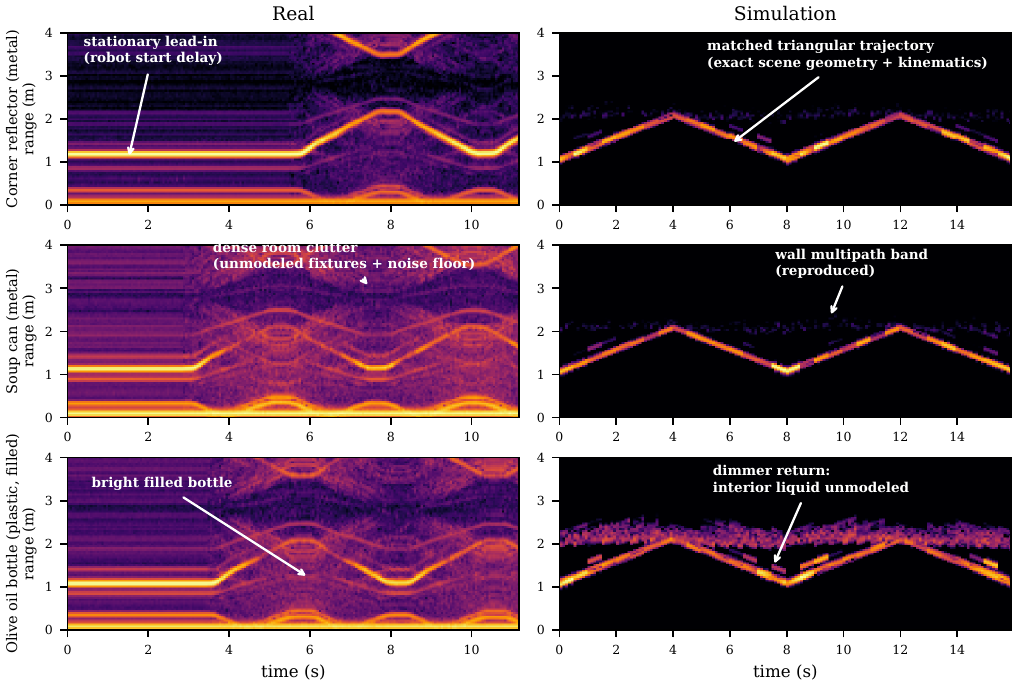}
  \caption{\textbf{Qualitative sim-vs-real fidelity} (rover depth sweeps, 1~m
start). Simulation reproduces the triangular target trajectory at correct
range and the brightness ordering across materials. Real recordings add a
stationary lead-in, denser clutter from unmodeled fixtures, and a brighter
filled bottle (interior liquid, Section~\ref{sec:future}).}
  \label{fig:qualitative}
\end{figure*}

\section{Object-Level Evaluation}
\label{sec:eval}

We evaluate the simulator on two complementary axes. \textbf{Fidelity}: does
it reproduce the signal structure a real radar measures? Qualitatively, do
simulated range-time signatures look like real ones (Fig.~\ref{fig:qualitative}),
and quantitatively, are the object-discriminative features preserved?
\textbf{Utility}: does a model trained on simulated data transfer to real
measurements (Section~\ref{sec:recog}, Section~\ref{sec:rover-recog})? We begin with
controlled turntable and linear-rail experiments showing the simulator
reproduces the two elementary motion transformations (rotation and
translation) and differentiates objects along the physical axes radar can
resolve.

\subsection{Setup}
\begin{figure*}[t]
\centering
\includegraphics[width=0.82\textwidth]{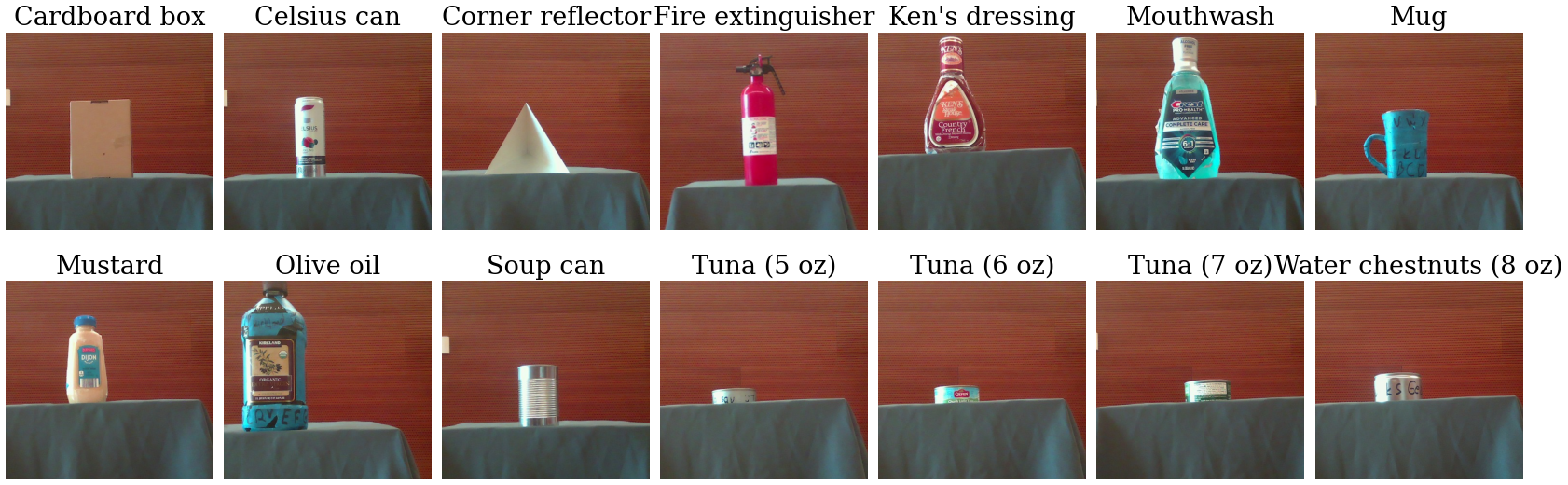}
\caption{The evaluated household objects, spanning the shape axis
(corner reflector, cylinders, tapered bottles) and the material axis (metal,
ceramic, plastic) independently.}
\label{fig:evaluated_objects}
\end{figure*}

Using the radar and simulator described in Section~\ref{sec:impl}, we evaluate objects chosen to span shape and material independently (Fig.~\ref{fig:evaluated_objects}). The shape axis ranges from a strong specular corner reflector, through rotationally symmetric cylinders, to asymmetric bottles and a mug-with-handle. The material axis spans three radar-relevant ITU-R P.2040 categories: metal, ceramic, and plastic. Metal objects include the corner reflector and several food and beverage cans; ceramic is represented by the mug; and plastic is represented by the olive oil, mouthwash, mustard, and Ken's dressing bottles. We include a near-identical food-can set to isolate size at fixed material and shape, since these objects differ primarily in height.

Each object is mounted on a programmable turntable rotating at one revolution per $\sim$25~s, placed at distances of 0.5, 1.0, 1.5, and 2.0~m, and recorded for 300~s, approximately twelve full rotations per recording, providing enough aspect-angle coverage to average out single-pose specular artifacts. We reconstruct the lab room from a phone scan with the same VLM-driven material
pipeline used at deployment (Section~\ref{sec:frame}), and generate matched
simulations whose object meshes, per-segment materials, object-radar
distance, rotation phase and rate, and IWR1843 chirp schedule are bit-aligned
with the real recordings. The same OBJ files serve as the simulator inputs and as the templates for
the physical reference objects (3D-printed or matched off-the-shelf items),
so any mesh inaccuracy matches what a deployment user authoring objects
from a phone scan would face.

\subsection{Disentangled Fidelity: Shape, Material, Size}
\label{sec:disent}

To understand what transfers from simulation to real radar, we isolate three object factors: shape, material, and size. Each factor is tested with a controlled contrast so that we can ask whether the simulator preserves the radar-observable structure rather than matching raw samples exactly.

\begin{figure*}[t]
\centering
\includegraphics[width=0.6\textwidth]{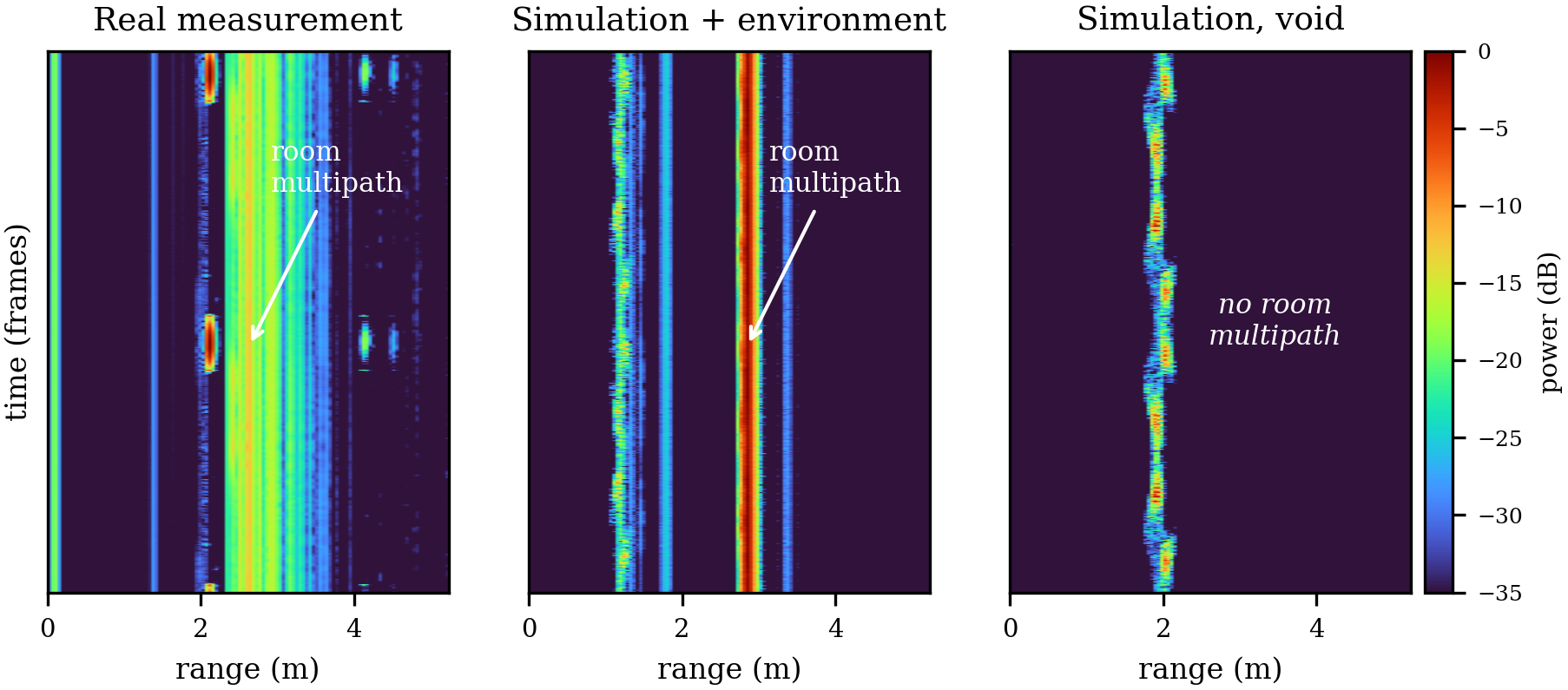}
\caption{\textbf{Indoor radar is shaped by the environment, not just the object.}
Range-time response of a rotating corner reflector at $2$~m: real (left),
simulation in the VLM-reconstructed room (middle), and an object-only void
simulation (right). The real return is dominated by room multipath
($\sim$75\% of its energy beyond the target), which the environment-aware
simulation recovers and the object-only simulation misses.}
\label{fig:env_ablation_rangetime}
\end{figure*}

\begin{figure*}[t]
\centering
\includegraphics[width=0.68\textwidth]{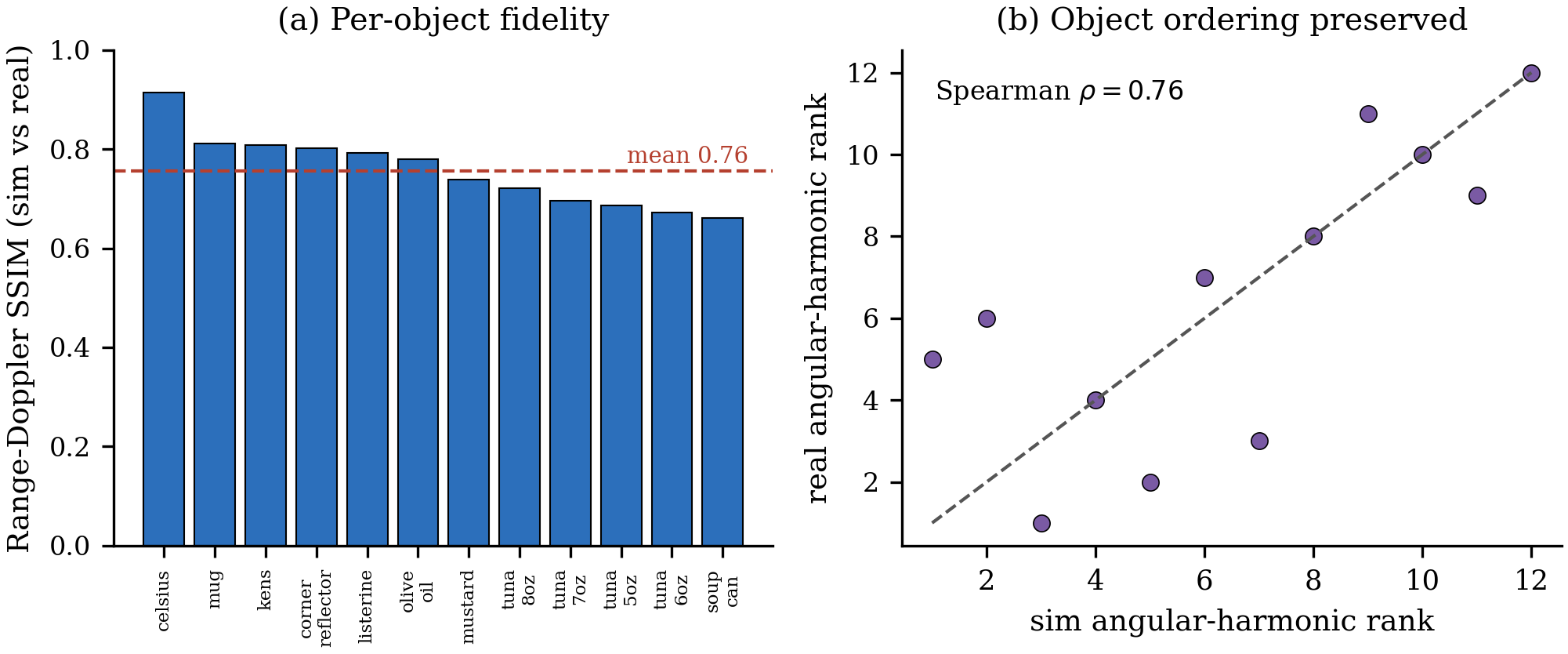}
\vspace{-0.5em}
\caption{\textbf{Turntable fidelity and structure preservation.}
(a) Per-object SSIM between simulated and real range-Doppler signatures
(mean $0.76$). (b) Objects ranked by shape signature in simulation and real
data. Simulation preserves the object ordering observed in reality
(Spearman $\rho=0.76$, Mantel $r{=}0.85$, $p{=}0.001$), matching the
structure recognition relies on.}
\label{fig:turntable_fidelity}
\vspace{-0.8em}
\end{figure*}

\noindent\textbf{Shape.}
A 77~GHz radar cannot directly resolve fine object geometry, but it can measure how the return changes with aspect angle. We therefore characterize shape by the angular modulation of the return during rotation. Symmetric objects, such as cylinders, produce nearly constant echoes, while faceted or asymmetric objects produce harmonic peaks when surfaces specularly align with the radar. We quantify this using the low-order angular harmonic ratio of the rotational reflectance envelope, computed from the 1--6 cycles/rotation band normalized by the DC component. This feature is invariant to absolute scale and distance. Real data shows the expected split: cans are flat, with angular harmonic ratios of $0.03$--$0.18$, while the corner reflector and bottles peak strongly at $0.72$--$2.5$. The simulator preserves this ordering, with sim-to-real rank correlation $0.89$ on cleanly localized objects and preserved inter-object structure (Mantel $r{=}0.85$, $p{=}0.001$, Fig.~\ref{fig:turntable_fidelity}). The signature is also stable with distance. The corner reflector has coefficient of variation $0.07$ across $0.6$--$2.1$~m. We restrict the metric to low-order harmonics because broadband modulation is contaminated by high-frequency facet ripple from the triangulated mesh ($\sim$128 cycles/rotation), which real smooth objects do not exhibit.

\noindent\textbf{Material.}
At matched distance and similar size, return brightness follows material class. Metal is brighter than plastic in both simulation and real data within the matched near-range group, showing that the ITU-driven material model captures the correct direction of the effect. The remaining errors are mostly absolute-scale effects: the simulator exaggerates the metal-plastic contrast and under-predicts the ceramic mug. These residuals are handled by the corner-reflector calibration in Section~\ref{sec:cr}, while the material ordering used by the recognizer is preserved.

\noindent\textbf{Size.}
The near-identical food-can set isolates size at fixed material and shape. Real radar cross section (RCS) does not increase consistently with can size. These objects differ mainly in height, but the 2-TX MIMO array is azimuth-only and has no elevation aperture, so the height differences are below what this configuration can resolve. The simulator sees the full mesh and can introduce a size-brightness trend that the real radar does not measure. We therefore treat fine object extent as below the physical resolution of this band and aperture, and characterize objects primarily by shape and material.

\subsection{Linear Transformations}
\label{sec:rail}

The turntable isolates rotation. We also test whether the simulator reproduces translational motion. A corner reflector is mounted on a $60$~cm motorized linear rail starting approximately $1$~m from the radar. We record two trajectories: radial motion toward and away from the radar, and diagonal motion at approximately $45^\circ$ to the line of sight. We simulate the same radial and radial-plus-lateral motions and compare the power-weighted peak-range trajectory against the real recordings.

Agreement is strong for radial motion. The measured range excursion is $0.50$~m in real data and $0.51$~m in simulation, with trajectory shape correlation $0.78$. For the diagonal trajectory, the expected radial excursion of a $60$~cm rail at approximately $45^\circ$ is approximately $0.42$~m, which the simulation matches at $0.44$~m. The real recording reads lower, at $0.22$~m, due to peak-tracking ambiguity in the multipath-cluttered short-range return. Together with the turntable result, this confirms that the simulator reproduces both rotation and translation in absolute range.

\subsection{Environment Ablation: Void vs.\ Reconstructed Scene}
\label{sec:env-ablation}
A central premise of \projectname{} is that the reconstructed environment, not
the object in isolation, must be simulated, because an indoor radar return is
shaped as much by the room as by the target. Walls, floor, and furniture create
multipath that arrives alongside the direct reflection. We test whether modeling
the environment actually makes the simulation more realistic by simulating the
same corner reflector two ways, in an empty void (object geometry only)
and inside the full VLM-reconstructed room, and asking which better matches the
real recording at each distance.

\begin{figure}[t]
\centering
\includegraphics[width=0.84\linewidth]{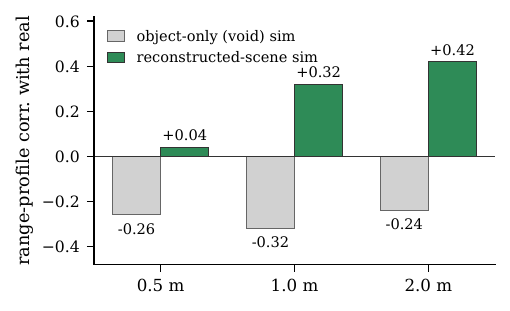}
\caption{\textbf{Simulating the room makes the simulation match
reality.} Correlation between simulated and real range profiles at each
turntable distance. The object-only simulation is anti-correlated with
real because it concentrates all energy at the target. The
reconstructed-scene simulation agrees increasingly well with range, as
room returns dominate the measurement.}
\label{fig:env_correlation}
\end{figure}

We compare the time-averaged range profile of each simulation against the real
corner-reflector recording
(Figs.~\ref{fig:env_ablation_rangetime} and \ref{fig:env_correlation}),
measured by Pearson correlation over range. Including the environment improves agreement at every
distance, and the gap widens with range. At $2$~m the correlation rises from
$-0.24$ (void) to $+0.42$ (reconstructed scene). The void simulation is
essentially uncorrelated or weakly anti-correlated with real
because it concentrates all energy at the target bin, whereas the real
profile is spread across the target and the room's reflections. The trend is intuitive: at short
range the strong direct return dominates and the two simulations look alike, but
as the object recedes the relative contribution of wall and floor multipath
grows, and only the environment-inclusive simulation reproduces it. This
confirms that environment-specific multipath is a first-order contributor to the
real signal, that the VLM-reconstructed scene supplies it, and, directly, that
including the environment makes the simulation more realistic.
We revisit this at deployment scale, where multipath is richer still, in
Section~\ref{sec:rover}.

\subsection{Corner-Reflector Radiometric Calibration}
\label{sec:cr}
Every analysis so far has compared structure rather than absolute
levels, because absolute radiometry does not survive the sim-to-real gap:
system gain and room-dependent multipath compression shift every level, and
the simulator over-separates material contrast. Tasks that need absolute
energy, material classification above all, require an anchor to the real
system. We adopt the cheapest available, a single trihedral corner
reflector (a
standard calibration target with known, aspect-stable cross-section) in
the deployment space, defining a per-feature affine map between simulated
and real energy-dependent features. This exceeds a scalar gain calibration
but uses no labels of any evaluated class. After anchoring, metal aligns
closely with real while dielectrics remain under-predicted, dominated by
liquid-filled containers (Section~\ref{sec:future}). The payoff is quantified
downstream. Material classification recovers from chance to $0.51$
(Section~\ref{sec:rover-recog}).

\subsection{Recognition with Simulation-Trained Representations}
\label{sec:recog}
The fidelity analyses establish that simulation preserves discriminative
structure. We now test whether that structure trains a recognizer. We train
the contrastive encoder of Section~\ref{sec:contrastive} exclusively on
simulated rotations ($30$ noise-augmented copies per object, matched to
measured SNR), 5 seeds. \textbf{(i) Labeled-probe}: real rotations are embedded by the frozen encoder
and classified by a leave-one-out nearest-class-centroid probe. The probe
itself is supervised ($n{-}1$ labeled rotations form the centroids,
$\sim$14 per object). What it isolates is the representation, namely the
encoder saw no real data, so the gap over the same probe on raw features is
attributable to simulation pre-training alone. The simulation-trained representation reaches
$\mathbf{95.3 \pm 1.5\%}$ 12-way recognition (chance $8.3\%$) versus
$89.7\%$ for the identical probe on raw features. Simulation pre-training contributes $+5.6$ points.
Since objects sit at different ranges, range could act as a class cue. A
range-stratified probe restricted to same-distance groups rules
this out: level and gap survive ($95.3 \pm 1.5\%$ vs.\ raw $90.3\%$,
stratified chance $24.8\%$). We also compare against an adapted published baseline. No existing
simulator natively supports object-centric indoor radar simulation, so we
adapt RF-Genesis~\cite{chen2023rfgenesis}, the closest published
generator, with best-effort modifications to run our task. These fixes
improve its performance, so the comparison is conservative. Under the
identical protocol, \projectname{} reaches $95.3 \pm 1.5\%$ versus
$90.1 \pm 1.5\%$ for the adapted baseline, a $+5.2$-point gap
attributable to our per-surface material physics and scene assembly, which
the baseline lacks.
\textbf{(ii) Label-free}: with no real
labels (unsupervised rank normalization only), fine-grained identity does
not transfer on the turntable, but attributes do. Geometry reaches $0.72$ vs.\ a
$0.58$ majority baseline, material $0.72$ vs.\ $0.67$. Label-free identity
is evaluated at deployment scale in Section~\ref{sec:rover-recog}.

\section{Contrastive Sim-Real Alignment}
\label{sec:contrastive}

The microbenchmarks show that simulated radar preserves object shape and material structure using hand-crafted features. We now learn this structure directly through a text-grounded representation and test whether it transfers across the sim-real gap without real labels. This applies contrastive alignment, previously used for human activity, to object and material recognition. \projectname{} enables the setting by producing physics-based signatures tied to object geometry and material.

We adapt vision-language contrastive learning~\cite{radford2021clip} to radar: a two-layer MLP encoder over the distance-controlled feature vector of Section~\ref{sec:disent} is aligned via InfoNCE~\cite{oord2018cpc} to short physical descriptions of each object (``metal, sharp single specular peak per rotation'', ``smooth round metal, steady''), embedded with MiniLM-L6-v2~\cite{wang2020minilm}. Training uses simulated signatures only. The text provides a shared physical anchor across domains (simulated and real signatures may differ in raw intensity, but both correspond to the same material-and-geometry description), and at test time real signatures are matched in the learned description space.

Trained on simulation and evaluated on real radar, the learned representation recovers the same physical structure observed in the microbenchmarks. Geometry is strongly preserved across the sim-real gap: a linear probe decodes symmetric-versus-asymmetric geometry from real embeddings at $85\%$ (chance $50\%$), and real signatures cluster by geometry class. Material structure transfers more weakly in the embedding ($40\%$ linear-probe accuracy against $33\%$ chance), consistent with the residual brightness gap the corner-reflector calibration addresses (Section~\ref{sec:cr}). Contrastive radar-text alignment produces an interpretable representation whose geometry axis transfers strongly to real measurements. It powers the recognition results of Section~\ref{sec:recog} and Section~\ref{sec:rover-recog}.

\section{Scene-Level Evaluation}
\label{sec:rover}

The microbenchmarks of Section~\ref{sec:eval} validate the simulator under
stripped-down conditions so each feature axis can be tested in isolation.
Deployment is harsher. A moving radar operates amid real environment
multipath, continuous platform motion, and the aspect-angle limits of a
single trajectory. This is the setting of Fig.~\ref{fig:motivation}, recognizing objects
with no real radar recordings from simulation-trained perception alone,
instantiated with a controlled household-object set measurable across
materials, shapes, and rooms.

We evaluate at the scene scale on multi-class object recognition under a
moving sensor in reconstructed deployment scenes
(Section~\ref{sec:rover-recog}), the direct scene-level analogue of the
recognition study in Section~\ref{sec:recog}, now under continuous platform
motion and real environment multipath. The downstream task is preceded by a
fidelity check (Section~\ref{sec:rover-fidelity}) and a disentanglement
analysis (Section~\ref{sec:rover-disent}) that isolate the object, distance,
and environment factors at the scene scale, so that any downstream success
or failure can be attributed to a specific axis of the simulator's behavior
rather than to an opaque end-to-end metric.

\begin{figure}[t]
  \centering
  \includegraphics[width=0.88\linewidth]{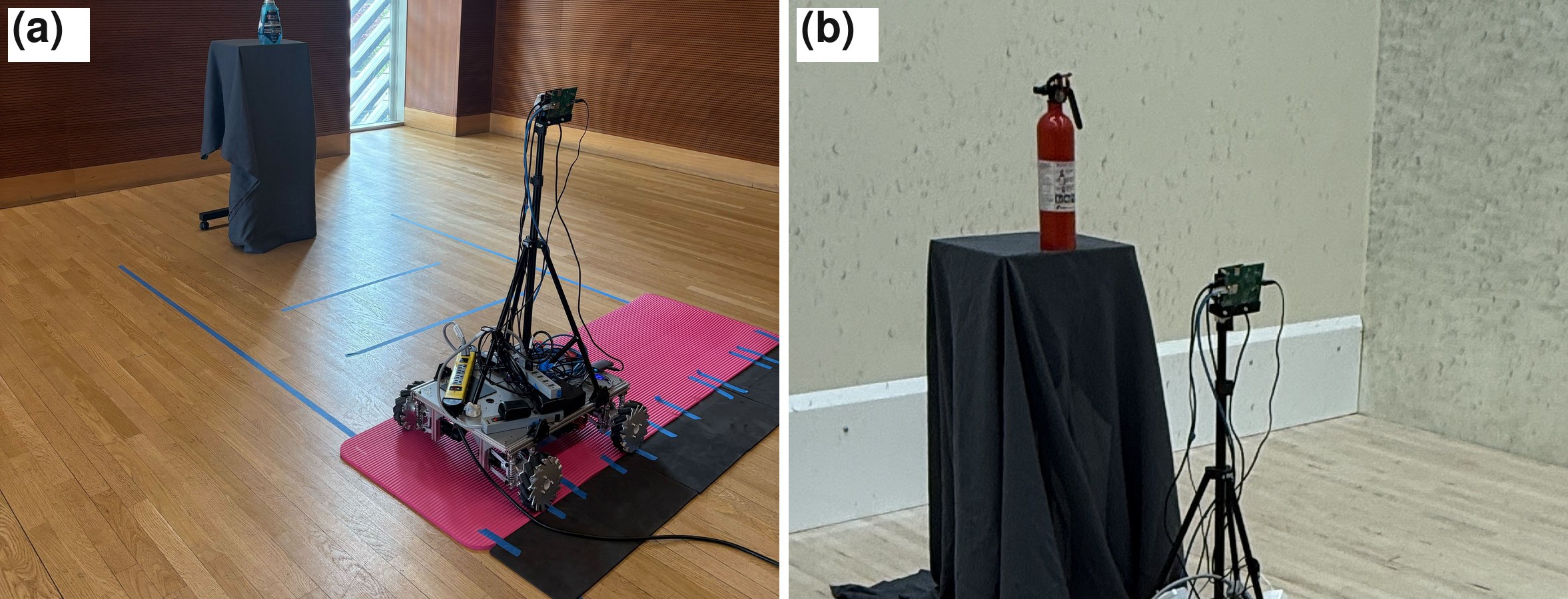}
  \caption{\textbf{Experimental setup.} The radar-equipped rover sweeps
  toward and past objects on a draped stand along taped trajectories at
  marked stand-offs: (a) furnished atrium (wood-paneled walls, glass
  doors). (b) enclosed squash court (concrete walls, wood floor). The
  contrasting multipath richness of the two rooms drives the
  environment-dependence analysis of Section~\ref{sec:rover-disent}.}
  \label{fig:setup_envs}
\end{figure}

\subsection{Setup}
Using the mobile robot platform described in Section~\ref{sec:impl}
(Fig.~\ref{fig:setup_envs}), the robot
performs repeated depth sweeps (toward/away from the object) and lateral
sweeps (past the object) at starting stand-offs of 0.5, 1.0, and 2.0~m for
each of 11 household objects, in two distinct environments: a furnished
atrium (glass doors, metal fixtures, plasterboard walls) and an enclosed
squash court (concrete walls, wood floor, minimal furniture), each
reconstructed with the VLM material pipeline. We generate matched \projectname{}
simulations along the recorded trajectories ($\sim$200 real recordings and
66 matched simulation runs per environment).

\noindent\textbf{Motion-specific feature design.}
\projectname{} is useful as a training prior when it preserves the object-discriminative features that each sensing motion exposes, not when it matches raw radar samples point by point. These features are determined by radar geometry. Rotation sweeps the full $360^\circ$ aspect angle and exposes shape through the low-order angular harmonic ratio of the rotation envelope. The rover motions expose different axes. \textbf{Depth motion} keeps aspect nearly fixed while range changes, so we use the tracked peak-range trajectory and the multipath share, measured as the fraction of energy outside the target gate ($\pm$3 range bins, $\pm$13~cm, over 0.3--5.5~m). \textbf{Lateral motion} sweeps a limited aspect range ($\sim$50--80$^\circ$), so we use the angular harmonic ratio of the pass-by envelope over that slice. Across all regimes, target-gate brightness provides the material feature and connects directly to the corner-reflector calibration of Section~\ref{sec:cr}. The contrastive representation of Section~\ref{sec:contrastive} is feature-agnostic and ingests the features supplied by each motion. Thus, the simulator's role is to reproduce the discriminative features physically exposed by a deployment trajectory.

\begin{figure}[t]
    \centering
    \includegraphics[width=0.88\linewidth]{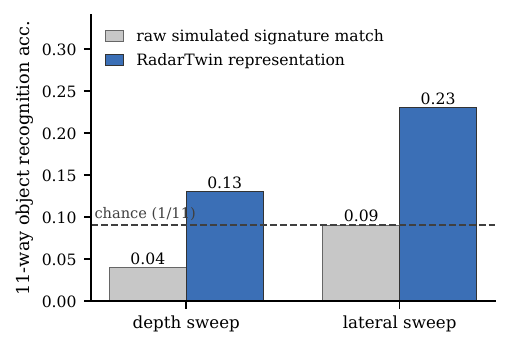}
    \caption{\textbf{Object recognition with no labeled real radar data from the evaluated classes}.
    Gray: matching raw simulated signatures
    fails. Blue: training on the \projectname{} representation reaches
    $2.5\times$ chance on the lateral sweep ($p<10^{-4}$).}
    \label{fig:alignment_helps}
\end{figure}

\subsection{Signal Fidelity Along Trajectories}
\label{sec:rover-fidelity}
Before asking whether the simulator transfers downstream, we check whether the simulated waveform tracks the real one as the rover moves through the room.
For every object-distance pair we time-align the simulated and real range-time
maps by cross-correlation of the target-bin energy profile, then characterize
the agreement on the three axes the depth motion physically resolves.
\textbf{Trajectory consistency} is the simulated peak-range curve as a function
of sensor position, scored by Pearson correlation against the real curve. This
is the radar analogue of comparing two moving point clouds on the same path:
if the sim has the right scene geometry and the rover's motion is faithfully
applied, the simulated and real triangular range-time signatures should
overlay. \textbf{Range-profile cross-correlation} averages the normalized range
profile over the recording and computes its alignment with the real profile.
This captures whether the sim places the target at the correct absolute range
and reproduces the relative ordering of secondary returns. \textbf{Doppler consistency} checks that the sim's chirp-to-chirp phase
progression at the target bin reproduces the recorded sweep kinematics
under depth motion.

Across the 11 objects, three distances, and two environments, the
results follow the pattern the physics predicts. Trajectory and range
placement are reproduced. The simulated triangular range-time signature
overlays the real sweep once per-recording start offsets are fitted, and
per-object mean range profiles correlate strongly for well-localized objects
(Pearson up to $+0.73$ for metals at matched distance). Absolute brightness
does not transfer. The simulator over-separates material contrast relative
to multipath-compressed real returns (a $\sim$25:1 ratio of separations in
log-energy), which is exactly the residual the corner-reflector calibration
of Section~\ref{sec:cr} anchors in deployment. What does transfer
reliably is ordering structure: of the per-feature sim-real Spearman
correlations across objects (distance-matched), significantly more are
positive than chance would allow in the depth regime ($78\%$ in the
atrium, $p{=}1.5{\times}10^{-5}$, $63\%$ in the court, $p{=}0.014$), with
the range-profile window, angular extent, and multipath share the most
consistent families, and the multipath-share scalar the single most
reliable feature across every room and motion regime. The metal$>$plastic
brightness ordering agrees in sign in $9$ of $9$ room$\times$distance
conditions. This is the deployment-scale analogue of
Section~\ref{sec:env-ablation}: the simulator is trustworthy about
relative structure (which object is brighter, wider, more
multipath-laden), and that is the level at which transfer succeeds.

\subsection{Disentangling Object, Distance, and Environment}
\label{sec:rover-disent}
At scene scale we cannot rotate the object through $360^\circ$. Instead we
hold two of \{object, distance, environment\} fixed and vary the
third, producing object, distance, and environment slices, so that any sim-real
disagreement on a slice is attributable to a single factor: the
per-object representation, the range-equation behaviour, or the scene
reconstruction respectively.

Two methodological findings from this analysis shape everything downstream.
First, distance is a confound. Pooling distances inflates sim-real
feature correlations (to $\sim$0.8) because both domains vary with range.
Distance-matched, per-feature correlations peak near $0.5$ and the
discriminative signal lives in the joint feature structure rather than any
single scalar. All scene-level transfer results are therefore distance-matched, and the
turntable probe is verified under range stratification (Section~\ref{sec:recog}). Second, the environment factor dominates: varying only the room,
label-free transfer succeeds in the multipath-rich squash court and remains
at chance in the open atrium (Section~\ref{sec:rover-recog}). The reconstruction
ablation below quantifies the same dependence within a single room.

\begin{table}[t]
\centering
\caption{Sim-real recognition by supervision level. Level (i) uses no real data of the evaluated
classes beyond unsupervised normalization. (ii) adds one unlabeled
corner-reflector recording. (iii) uses a handful of labeled rotations per
class. $^{\dagger}$For 3-way shape, chance is the majority-class baseline
rather than uniform $1/3$. }
\label{tab:rover_recog}
\small
\setlength{\tabcolsep}{4pt}
\begin{tabular}{llccc}
\toprule
Level & Setting & Raw & \projectname{} & Chance \\
\midrule
(i)   & Rover lateral, 11-way obj.    & 0.09 & \textbf{0.23}\,\tiny$\pm$.02 & 0.09 \\
(i)   & Rover depth, 11-way obj.      & 0.04 & 0.13\,\tiny$\pm$.02 & 0.09 \\
(i)   & Rover lateral, 3-way shape    & 0.55 & \textbf{0.65}\,\tiny$\pm$.01 & 0.55$^{\dagger}$ \\
(ii)  & Rover depth, 3-way material   & 0.35 & \textbf{0.51} & 0.33 \\
(ii)  & Rover lateral, 3-way material & 0.30 & 0.48 & 0.33 \\
(iii) & Turntable, 12-way obj.        & 0.90 & \textbf{0.95}\,\tiny$\pm$.02 & 0.08 \\
\bottomrule
\end{tabular}
\end{table}

\subsection{Scene-Level Recognition Across Supervision Levels}
\label{sec:rover-recog}
We evaluate $11$-way object recognition with the simulation-trained pipeline
at three clearly separated supervision levels. All numbers are means over
5 seeds with recording-level majority voting (Fig.~\ref{fig:alignment_helps},
Table~\ref{tab:rover_recog}).

\noindent\textbf{(i) Label-free.} Training on simulation alone, with no
real data of the evaluated classes beyond unsupervised per-feature rank normalization (transductive: it uses
the unlabeled evaluation recordings' feature ranks but no labels), a random forest trained on rank-normalized simulated frames and
majority-voted per recording reaches, in the squash court,
$0.23 \pm 0.02$ on the lateral sweep ($2.5\times$ chance $0.091$,
binomial $p<10^{-4}$ over recordings) and $0.13 \pm 0.02$ on the depth
sweep. The unaligned raw-signature baselines reach only $0.09$ and $0.04$
(Fig.~\ref{fig:alignment_helps}). The raw whole-signature match does not transfer. The
aligned representation does. Shape classification on the lateral sweep
reaches $0.65 \pm 0.01$ (majority baseline $0.55$).

\begin{figure}[t]
\centering
\includegraphics[width=0.84\linewidth]{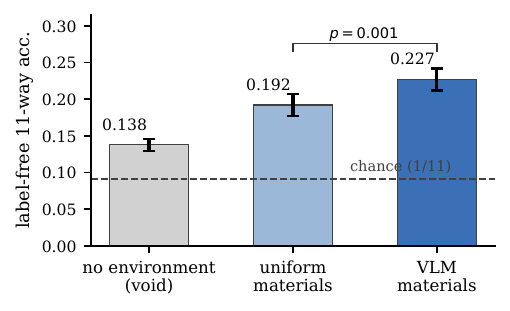}
\caption{\textbf{The reconstruction is what makes unseen-object
recognition work.} Label-free 11-way accuracy (lateral sweep, means and
std over 5 seeds) as the simulation is degraded. Removing VLM materials
costs accuracy ($p{=}0.001$, paired test over recordings), and
removing the environment drops recognition to near chance.}
\label{fig:recon_ablation}
\end{figure}

\noindent\textbf{(ii) One calibration recording.} Adding the single
corner-reflector recording of Section~\ref{sec:cr} (no labels for any
evaluated class) recovers material classification to $0.51 \pm 0.00$ (depth) and
$0.48 \pm 0.03$ (lateral) against $0.33$ chance over 5 seeds, evaluated
with the calibration object excluded.

\noindent\textbf{(iii) Few labeled examples.} With a handful of labeled
real rotations per class, the simulation-pre-trained representation reaches
$95.3 \pm 1.5\%$ 12-way recognition in the controlled setting
(Section~\ref{sec:recog}), $+5.6$ points over the identical probe on raw
features.

\noindent\textbf{Reconstruction ablation.} To isolate what the
material-aware reconstruction contributes, we regenerate all lateral
simulations under two degraded conditions and rerun the identical label-free
protocol: uniform (room geometry kept, every surface forced to
plasterboard) and void (object and stand only)
(Fig.~\ref{fig:recon_ablation}). Recognition degrades monotonically: VLM materials $0.227 \pm 0.015$,
uniform $0.192 \pm 0.015$, void $0.138 \pm 0.008$ (void not above
chance). The VLM-over-uniform margin is significant. On paired
per-recording decisions, VLM is uniquely correct on $23$ recordings versus
$6$ for uniform ($p{=}0.001$, paired exact test), and it leads in all five
seeds. The
environment carries the largest share of the transferable signal and
per-surface VLM materials add a further margin. Shape is material-invariant
($0.66$ uniform vs.\ $0.65$ VLM) but collapses without the environment
($0.54$). Both stages of the reconstruction earn their place.

The comparison that matters for an unseen class is against the no-data
alternative, exactly the deployment case of Fig.~\ref{fig:motivation}:
raw signature matching fails outright, while the aligned
representation recovers significant recognition. Simulation provides a
usable prior exactly where real data is unavailable, and each increment of
deployment effort purchases a measured increase in capability.

\section{Discussion}

\noindent\textbf{What transfers, and why it is useful.} The simulator
reproduces an object's distance-invariant shape signature and material
class even though raw signals differ substantially. Simulated data is
imperfect in appearance but right about the features that matter, so a
recognizer trained on it generalizes when real data is limited.

\noindent\textbf{Match simulation complexity to where the signal lives.}
On the turntable, identity is carried by object-intrinsic aspect modulation
and object-only simulation suffices. At deployment scale the moving sensor
couples the object to the room, environment reconstruction becomes a hard
requirement (Section~\ref{sec:env-ablation}), and transfer tracks environment
richness.

\section{Limitations and Future Work}
\label{sec:future}

Experiments ran in clear visual conditions. The smoke-filled deployments
of Fig.~\ref{fig:motivation} are unaffected by this choice, since radar is
visibility-invariant and the camera serves only trajectory matching. The
fixed protocol makes comparison with further published simulators
mechanical, extending the adapted RF-Genesis baseline of
Section~\ref{sec:recog}. Several simulator fidelity gaps are concrete targets for future work: the
under-modeling of dielectric reflectivity (most notably the liquid-filled
containers below) and the facet-ripple artifact from the triangulated
mesh. On the method
side, the absolute domain gap restricts label-free transfer to
ordering-based methods (Section~\ref{sec:rover-recog}). A natural extension is a
paired sim-real contrastive term on top of the text-grounded alignment of
Section~\ref{sec:contrastive}. Transfer is also environment-dependent (it
succeeded in the multipath-rich court but not the open atrium).
Characterizing which environment properties predict transfer is open, and
the evaluation spans two rooms and twelve objects. Scaling to more
environments and object categories is the clearest path to strengthening
the deployment claim.

Several of our objects are filled plastic bottles. At 77~GHz the
dominant return comes from the high-permittivity liquid contents, which our
surface-material pipeline does not model, leaving these returns too dim
even after calibration. Faithful simulation requires modeling contents as a
distinct dielectric volume. More broadly, the VLM material stage should
reason about an object's interior, not only its visible surface.

\section{Conclusion}
\looseness=-1 We presented \projectname{}, which generates mmWave radar
training data from a commodity 3D scan of the deployment space via VLM-inferred
per-surface materials and physics-based multi-bounce FMCW simulation. Its
central methodological claim is that simulator fidelity should be evaluated
at the level of the object-discriminative features a perception model
consumes, not at the raw-signal level. The central capability this enables
is recognition trained on simulation alone, before any labeled radar data
exists at the deployment site: because simulated and real radar share the
same object-discriminative features, a representation trained on simulation
alone---with no real labels---recognizes real objects at $2.5\times$ chance
($p<10^{-4}$). Simulation provides a usable training prior exactly where real
data is unavailable, and additional real supervision sharpens this prior
rather than being required for it: a single unlabeled
corner-reflector capture recovers material classification from chance to
$0.51$, and a handful of labeled rotations lifts the same representation to
$95.3 \pm 1.5\%$ 12-way recognition. Our end-to-end framework and paired
dataset let others extend this analysis to new environments, objects, and
radar configurations.

\begin{acks}
This research was partially supported by COGNISENSE, one of seven
centers in JUMP 2.0, a Semiconductor Research Corporation (SRC) program
sponsored by DARPA, as well as the National Science Foundation under
Grant Number CNS-1943396. The views and conclusions contained here are
those of the authors and should not be interpreted as necessarily
representing the official policies or endorsements, either expressed or
implied, of Columbia University, NSF, SRC, DARPA, or the U.S. Government
or any of its agencies.
\end{acks}
\bibliographystyle{ACM-Reference-Format}
\bibliography{references}

@inproceedings{chen2023rfgenesis,
  author    = {Chen, Xingyu and Zhang, Xinyu},
  title     = {{RF Genesis}: Zero-Shot Generalization of mmWave Sensing through Simulation-Based Data Synthesis and Generative Diffusion Models},
  booktitle = {Proceedings of the 21st ACM Conference on Embedded Networked Sensor Systems (SenSys '23)},
  year      = {2023},
  publisher = {ACM},
  doi       = {10.1145/3625687.3625798}
}

@inproceedings{cao2024mmclip,
  author    = {Cao, Qiming and Xue, Hongfei and Liu, Tianci and Wang, Xingchen and Wang, Haoyu and Zhang, Xincheng and Su, Lu},
  title     = {{mmCLIP}: Boosting mmWave-based Zero-shot {HAR} via Signal-Text Alignment},
  booktitle = {Proceedings of the 22nd ACM Conference on Embedded Networked Sensor Systems (SenSys '24)},
  pages     = {184--197},
  year      = {2024},
  publisher = {ACM},
  doi       = {10.1145/3666025.3699331}
}

@inproceedings{chen2024rfcanvas,
  author    = {Chen, Xingyu and Zhang, Xinyu},
  title     = {{RFCanvas}: Modeling {RF} Channel by Fusing Visual Priors and Few-shot {RF} Measurements},
  booktitle = {Proceedings of the 22nd ACM Conference on Embedded Networked Sensor Systems (SenSys '24)},
  year      = {2024},
  publisher = {ACM},
  doi       = {10.1145/3666025.3699351}
}

@article{schoffmann2021vira,
  author  = {Sch{\"o}ffmann, Christian and Ubezio, Barnaba and B{\"o}hm, Christoph and M{\"u}hlbacher-Karrer, Stephan and Zangl, Hubert},
  title   = {Virtual Radar: Real-Time Millimeter-Wave Radar Sensor Simulation for Perception-Driven Robotics},
  journal = {IEEE Robotics and Automation Letters},
  volume  = {6},
  number  = {3},
  pages   = {4704--4711},
  year    = {2021},
  doi     = {10.1109/LRA.2021.3068916}
}

@article{shenron2024,
  author  = {Bansal, Kshitiz and Reddy, Gautham and Bharadia, Dinesh},
  title   = {{SHENRON} -- Scalable, High Fidelity and Efficient Radar Simulation},
  journal = {IEEE Robotics and Automation Letters},
  volume  = {9},
  number  = {2},
  pages   = {1644--1651},
  year    = {2024},
  doi     = {10.1109/LRA.2023.3343168}
}

@inproceedings{mishra2025cshenron,
  author    = {Mishra, Pushkal and Srivastava, Satyam and Li, Jerry and Bansal, Kshitiz and Bharadia, Dinesh},
  title     = {Demo Abstract: {C-Shenron}: A Realistic Radar Simulation Framework for {CARLA}},
  booktitle = {Proceedings of the 23rd ACM Conference on Embedded Networked Sensor Systems (SenSys '25)},
  pages     = {726--727},
  year      = {2025},
  publisher = {ACM},
  doi       = {10.1145/3715014.3724379}
}

@inproceedings{he2023fusang,
  author    = {He, Guorong and Chen, Shichao and Xu, Dongheng and Chen, Xianbin and Xie, Yang and Wang, Xiaohua and Fang, Dingyi},
  title     = {Fusang: Graph-inspired Robust and Accurate Object Recognition on Commodity mmWave Devices},
  booktitle = {Proceedings of the 21st Annual International Conference on Mobile Systems, Applications and Services (MobiSys '23)},
  pages     = {489--502},
  year      = {2023},
  publisher = {ACM}
}

@inproceedings{chi2024rfdiffusion,
  author    = {Chi, Guoxuan and Yang, Zheng and Wu, Chenshu and Xu, Jingao and Gao, Yuchong and Liu, Yunhao and He, Tianyue},
  title     = {{RF-Diffusion}: Radio Signal Generation via Time-Frequency Diffusion},
  booktitle = {Proceedings of the 30th Annual International Conference on Mobile Computing and Networking (MobiCom '24)},
  pages     = {77--92},
  year      = {2024},
  publisher = {ACM}
}

@inproceedings{dodds2025nlos,
  author    = {Dodds, Laura and Boroushaki, Tara and Zhou, Kaichen and Adib, Fadel},
  title     = {Non-Line-of-Sight {3D} Object Reconstruction via mmWave Surface Normal Estimation},
  booktitle = {Proceedings of the 23rd Annual International Conference on Mobile Systems, Applications and Services (MobiSys '25)},
  pages     = {445--458},
  year      = {2025},
  publisher = {ACM}
}

@inproceedings{chow2025physbench,
  author    = {Chow, Wei and Mao, Jiageng and Li, Boyi and Seita, Daniel and Guizilini, Vitor and Wang, Yue},
  title     = {{PhysBench}: Benchmarking and Enhancing Vision-Language Models for Physical World Understanding},
  booktitle = {International Conference on Learning Representations (ICLR)},
  year      = {2025}
}

@inproceedings{bell2015minc,
  author    = {Bell, Sean and Upchurch, Paul and Snavely, Noah and Bala, Kavita},
  title     = {Material Recognition in the Wild with the Materials in Context Database},
  booktitle = {Proceedings of the IEEE Conference on Computer Vision and Pattern Recognition (CVPR)},
  year      = {2015}
}

@inproceedings{radford2021clip,
  author    = {Radford, Alec and Kim, Jong Wook and Hallacy, Chris and Ramesh, Aditya and Goh, Gabriel and Agarwal, Sandhini and Sastry, Girish and Askell, Amanda and Mishkin, Pamela and Clark, Jack and Krueger, Gretchen and Sutskever, Ilya},
  title     = {Learning Transferable Visual Models From Natural Language Supervision},
  booktitle = {Proceedings of the 38th International Conference on Machine Learning (ICML)},
  year      = {2021}
}

@article{oord2018cpc,
  author  = {van den Oord, Aaron and Li, Yazhe and Vinyals, Oriol},
  title   = {Representation Learning with Contrastive Predictive Coding},
  journal = {arXiv preprint arXiv:1807.03748},
  year    = {2018}
}

@inproceedings{wang2020minilm,
  author    = {Wang, Wenhui and Wei, Furu and Dong, Li and Bao, Hangbo and Yang, Nan and Zhou, Ming},
  title     = {{MiniLM}: Deep Self-Attention Distillation for Task-Agnostic Compression of Pre-Trained Transformers},
  booktitle = {Advances in Neural Information Processing Systems (NeurIPS)},
  year      = {2020}
}

@article{hoydis2023sionna,
  author  = {Hoydis, Jakob and Aoudia, Faycal Ait and Cammerer, Sebastian and Nimier-David, Merlin and Maggi, Lorenzo and Marcus, Guillermo and Keller, Alexander},
  title   = {{Sionna RT}: Differentiable Ray Tracing for Radio Propagation Modeling},
  journal = {arXiv preprint arXiv:2303.11103},
  year    = {2023}
}

@misc{jakob2022mitsuba3,
  author = {Jakob, Wenzel and Speierer, S{\'e}bastien and Roussel, Nicolas and Nimier-David, Merlin and Vicini, Delio and Zeltner, Tizian and Nicolet, Baptiste and Crespo, Miguel and Leroy, Vincent and Zhang, Ziyi},
  title  = {{Mitsuba 3} Renderer},
  year   = {2022},
  note   = {Version 3.x, \url{https://mitsuba-renderer.org}}
}

@techreport{itu2040,
  author      = {{International Telecommunication Union}},
  title       = {Recommendation {ITU-R P.2040-4}: Effects of Building Materials and Structures on Radiowave Propagation above about 100 {MHz}},
  institution = {ITU-R},
  year        = {2023}
}

@misc{matlab_radar,
  author = {{The MathWorks, Inc.}},
  title  = {Radar Toolbox},
  year   = {2024},
  note   = {\url{https://www.mathworks.com/products/radar.html}}
}

@misc{ansys_hfss,
  author = {{Ansys, Inc.}},
  title  = {{Ansys HFSS}: 3D High Frequency Simulation Software},
  year   = {2024},
  note   = {\url{https://www.ansys.com/products/electronics/ansys-hfss}}
}

@misc{remcom,
  author = {{Remcom, Inc.}},
  title  = {{Wireless InSite} Propagation Software},
  year   = {2024},
  note   = {\url{https://www.remcom.com/wireless-insite-em-propagation-software}}
}

@article{chen2024internvl,
  author  = {Chen, Zhe and Wang, Weiyun and Tian, Hao and Ye, Shenglong and Gao, Zhangwei and Cui, Erfei and Tong, Wenwen and others},
  title   = {Expanding Performance Boundaries of Open-Source Multimodal Models with Model, Data, and Test-Time Scaling ({InternVL 2.5})},
  journal = {arXiv preprint arXiv:2412.05271},
  year    = {2024}
}

@inproceedings{ahuja2021vid2doppler,
  author    = {Ahuja, Karan and Jiang, Yue and Goel, Mayank and Harrison, Chris},
  title     = {{Vid2Doppler}: Synthesizing Doppler Radar Data from Videos for Training Privacy-Preserving Activity Recognition},
  booktitle = {Proceedings of the 2021 CHI Conference on Human Factors in Computing Systems (CHI '21)},
  year      = {2021},
  publisher = {ACM}
}

@inproceedings{lu2020millimap,
  author    = {Lu, Chris Xiaoxuan and Rosa, Stefano and Zhao, Peijun and Wang, Bing and Chen, Changhao and Stankovic, John A. and Trigoni, Niki and Markham, Andrew},
  title     = {See Through Smoke: Robust Indoor Mapping with Low-cost mmWave Radar},
  booktitle = {Proceedings of the 18th International Conference on Mobile Systems, Applications, and Services (MobiSys '20)},
  year      = {2020}
}

@inproceedings{lu2024radarize,
  author    = {Sie, Emerson and Wu, Xinyu and Guo, Heyu and Vasisht, Deepak},
  title     = {Radarize: Enhancing Radar {SLAM} with Generalizable Doppler-Based Odometry},
  booktitle = {Proceedings of the 22nd Annual International Conference on Mobile Systems, Applications and Services (MobiSys '24)},
  year      = {2024}
}

@article{lam2025minav,
  author    = {Lam, Maisy and Herrera, Joshua and Afzal, Sayed Saad and Zhou, Kaichen and Adib, Fadel},
  title     = {MiNav: Autonomous Drone Navigation Indoors Using Millimeter-Waves},
  journal   = {Proceedings of the ACM on Interactive, Mobile, Wearable and Ubiquitous Technologies},
  year      = {2025},
  doi       = {10.1145/3749464}
}

@article{ling2024uranus,
  author    = {Ling, Kai and Zhao, Running and others},
  title     = {Uranus: Empowering Generalized Gesture Recognition with Mobility through Generating Large-scale mmWave Radar Data},
  journal   = {Proceedings of the ACM on Interactive, Mobile, Wearable and Ubiquitous Technologies},
  volume    = {8},
  number    = {4},
  year      = {2024},
  doi       = {10.1145/3699754}
}

@inproceedings{duan2025ifr,
  author    = {Duan, K. and Zhu, Z. and Zou, Z.},
  title     = {Indoor {FireRescue} Radar: {4D} Indoor Millimeter Wave Dataset and Analysis for Hazardous Environment Perception},
  booktitle = {Proceedings of the IEEE/RSJ International Conference on Intelligent Robots and Systems (IROS)},
  pages     = {18620--18627},
  year      = {2025}
}

@inproceedings{wang2025moge,
  author    = {Wang, Ruicheng and Xu, Sicheng and Dai, Cassie and Xiang, Jianfeng and Deng, Yu and Tong, Xin and Yang, Jiaolong},
  title     = {{MoGe}: Unlocking Accurate Monocular Geometry Estimation for Open-Domain Images with Optimal Training Supervision},
  booktitle = {Proceedings of the IEEE/CVF Conference on Computer Vision and Pattern Recognition (CVPR)},
  year      = {2025}
}
\end{document}